\documentclass{article}
\usepackage[utf8]{inputenc}
\usepackage{tikz}

\usepackage{csquotes}
\usepackage{paralist}
\usepackage{amsmath}
\usepackage{amssymb}
\usepackage{graphicx}
\usepackage{verbatim}
\usepackage{natbib}

\usetikzlibrary{shapes}
\usetikzlibrary{arrows.meta,backgrounds,automata}
\usetikzlibrary{positioning,shapes,matrix,calc}
\tikzstyle{vertex}=[circle, draw, fill=gray!80!white,thick,scale=1.2]
\tikzstyle{edge}=[draw=black, thick,-]

\usetikzlibrary{calc,shapes, decorations.pathmorphing}

\title{\bf Neural Algorithmic Reasoning}
\author{Petar Veli\v{c}kovi\'{c} and Charles Blundell}
\date{DeepMind}

\begin{document}

\maketitle

\begin{abstract}
\noindent Algorithms have been fundamental to recent global technological advances and, in particular, they have been the cornerstone of technical advances in one field rapidly being applied to another. We argue that algorithms possess fundamentally different qualities to deep learning methods, and this strongly suggests that, were deep learning methods better able to mimic algorithms, generalisation of the sort seen with algorithms would become possible with deep learning---something far out of the reach of current machine learning methods. Furthermore, by representing elements in a continuous space of learnt algorithms, neural networks are able to adapt known algorithms more closely to real-world problems, potentially finding more efficient and pragmatic solutions than those proposed by human computer scientists.

Here we present neural algorithmic reasoning---the art of building neural networks that are able to execute algorithmic computation---and provide our opinion on its transformative potential for running classical algorithms on inputs previously considered inaccessible to them.
\end{abstract}

\paragraph{Algorithms and Deep Learning}

Algorithms are pervasive in modern society---from elevators, microwave ovens and other household equipment to procedures for electing government officials.
Algorithms allow us to automate and engineer systems that reason.
Remarkably, algorithms applied in one domain---such as a microwave oven---may be slightly adjusted and deployed in a completely different domain---such as a heart pacemaker (e.g., a control algorithm such as PID). That is not to say that you would expect to be able to safely run a microwave oven using a pacemaker (or vice versa) without modification, but the same recipe underlies both constructions.

An undergraduate textbook on algorithms \citep{cormen2009introduction} will cover fewer than 60 distinct algorithms.
A subset of these will serve as the useful basis for someone's life-long career in software engineering in almost any domain.
Part of the skill of a software engineer lies in choosing which algorithm to use, when, and in combination with what else.
Only rarely will an entirely novel algorithm be warranted.

This same algorithmic basis could also help us solve one of the hardest problems in deep learning: generalisation.
Deep learning methods learn from data and are then deployed to make predictions or decisions.
The core generalisation concern is: will it work in a new situation? In other words, from training data, will the deep learning method \emph{generalise} to the new situation.
Under certain assumptions, guarantees can be given but so far these are in simple cases.

Algorithms, on the other hand, typically come with strong general guarantees. The invariances of an algorithm can be stated as a precondition and a postcondition, combined with how the time and space complexity scales with input size.
The precondition states what the algorithm will assume is true about its inputs, and the post-condition will state what the algorithm can then guarantee about its outputs after its execution.
For example, the precondition of a sorting algorithm may specify what kind of input it expects (e.g., a finite list of integers allocated in memory it can modify) and then the postcondition might state that after execution, the input memory location contains the same integers but in ascending order.

Even with something as elementary as sorting, neural networks cannot provide guarantees of this kind: neural networks can be demonstrated to work on certain problem instances and to generalise to certain larger instances than were in the training data. There is no guarantee they will work for all problem sizes, unlike good sorting algorithms, nor even on all inputs of a certain size.

Algorithms and the predictions or decisions learnt by deep learning have very different properties---the former provide strong guarantees but are inflexible to the problem being tackled, whilst the latter provide few guarantees but can adapt to a wide range of problems.
Understandably, work has considered how to get the best of both.
Induction of algorithms from data will have significant implications in computer science: better approximations to intractable problems, previously intractable problems shown to be tractable in practice, and algorithms that can be optimised directly for the hardware that is executing them with little or no human intervention.

Already several approaches have been explored for combining deep learning and algorithms.
Inspired by deep reinforcement learning, deep learning methods can be trained to use existing, known algorithms as fixed external tools \citep{reed2015neural,li2020strong}. This very promising approach works well---when the existing known algorithms fit the problem at hand.
This is somewhat reminiscent of the software engineer wiring together a collection of known algorithms.
An alternative approach is to teach deep neural networks to imitate the workings of an existing algorithm, by producing the same output, and in the strongest case by replicating the same intermediate steps \citep{graves2014neural,kaiser2015neural,kurach2015neural,velivckovic2020pointer}.
In this form, the algorithm itself is encoded directly into the neural network before it is executed.
This more fluid representation of the algorithm allows learning to adapt the internal mechanisms of the algorithm itself via feedback from data.
Furthermore, a single network may be taught multiple known algorithms and abstract commonalities among them  \citep{velivckovic2019neural}, allowing novel algorithms to be derived.
Both of these approaches build atop known algorithms. In the former case, new combinations of existing algorithms can be learnt. Excitingly, in the latter case, new variants or adaptations of algorithms can be learnt, as the deep neural network is more malleable than the original algorithm.

At present, in computer science, a real world problem is solved by first fitting the problem to a known class of problems (such as sorting all numbers), and then an appropriate algorithm chosen for this known problem class.
This known problem class may actually be larger than that exhibited by the real world problem, and so the chosen algorithm may be suboptimal in practice (for example, the known problem class may be \textsf{NP-hard}, but all real world examples are actually in \textsf{P}, so can be solved in polynomial time).
Instead, by combining deep learning and algorithms together, an algorithm can be fit directly to the real world problem, without the need for the intermediate proxy problem.
 
\paragraph{Algorithms in the real world}
To elaborate on how neural networks can more directly be fused with algorithmic computation, we will take a step back and consider the theoretical  motivation for designing algorithms.

Algorithms can represent the purest form of problem-solving. The Church-Turing thesis states that a problem of any kind is computationally solvable if and only if there exists an algorithm that solves it when executed on the model of computation known as a Turing machine. Thus, solvability of problems necessitates existence of suitable algorithms for them.

Algorithms reason about problems in an \emph{abstract} space, where the inputs conform to stringent \emph{pre-conditions}. Under this lens, it becomes far easier to guarantee correctness (in the form of stringent \emph{post-conditions}), provide performance guarantees, support elegant and interpretable pseudocode, and perhaps most importantly, draw clear connections between problems that may be otherwise hard to relate. 

The theoretical utility of algorithms is unfortunately at timeless odds with the practical motivation for designing them: to apply them to \emph{real-world} problems. Clear examples of both the appeal of algorithmic reasoning and the apparent dissonance it has with practical applications were known as early as 1955---within a write-up from Harris and Ross \citep{harris1955fundamentals}, which studied the bottleneck properties of railway networks. 

By studying the problem in an abstract space (railway junctions being nodes in a graph, and edges between them endowed with scalar capacities, specifying the limits of traffic flow along edges), the authors formalised the bottleneck finding task as a \emph{minimum-cut} problem. Observing the problem in this abstract space made it easily relatable to the (otherwise seemingly unrelated) \emph{maximum-flow} problem. In fact, studying the problem under this lens not only enabled a strong theoretical connection between these two problems, it also spearheaded decades of research into efficient algorithms in flow networks.

However, there is a fundamental limitation to this kind of abstraction. In order for all of the above to be applicable, we need to ``compress'' all of the complexity of the real-world railway network into single-scalar capacities for every edge. And, as the authors themselves remark: \begin{displayquote}
	\emph{``The evaluation of both railway system and individual track capacities is, to a considerable extent, an art. The authors know of no tested mathematical model or formula that includes all of the variations and imponderables that must be weighed.* Even when the individual has been closely associated with the particular territory he is evaluating, the final answer, however accurate, is largely one of judgment and experience.''}
\end{displayquote}
This remark may be of little importance to the theoretical computer scientist, but it has strong implications on applying classical algorithms on \emph{natural} inputs, that hold to this day. If data is manually converted from raw to abstract form, this often implies drastic information loss, making our problem no longer accurately portray the dynamics of the real world. Hence the algorithm will give a \emph{perfect} solution, but in a potentially \emph{useless} setup. Even more fundamentally, the data we need to apply the algorithm may be only \emph{partially observable}---in which case, the algorithm could even be rendered inapplicable.

In order to circumvent this issue, we may recall that the ``deep learning revolution'' occurred with neural networks replacing the use of manual feature extractors from raw data, causing significant gains in performance. Accordingly, as our issues stem from manually converting complex natural inputs to algorithmic inputs, we propose applying neural networks in this setting as well.

\paragraph{Algorithmic bottlenecks and neural algorithm execution}

Directly predicting the algorithmic inputs from raw data often gives rise to a very peculiar kind of \emph{bottleneck}. Namely, the richness of the real world (e.g. noisy real-time traffic data) still needs to be compressed into scalar values (e.g. edge weights in a path-finding problem). The algorithmic solver then commits to using these values and assumes they are free of error---hence, if we don't have sufficient data to estimate these scalars properly, the resulting environment where the algorithm is executed does not accurately portray the real-world problem, and results may be suboptimal, especially for low data setups.

To break the algorithmic bottleneck, it would be preferential to have our neural network consistently producing \emph{high-dimensional} representations. This means that the computations of our algorithm also must be made to operate over high-dimensional spaces. The most straightforward way to achieve this is  replacing the algorithm itself with a neural network---one which mimics the algorithm's operations in this latent space, such that the desirable outputs are decodable from those latents. The recently resurging area of algorithmic reasoning \citep[Section 3.3.]{cappart2021combinatorial} exactly studies the ways in which such algorithmically-inspired neural networks can be built, primarily through learning to execute the algorithm from abstractified inputs.

Algorithmic reasoning provides methods to train useful \emph{processor networks}, such that within their parameters we find a combinatorial algorithm that is
\begin{inparaenum}
	\item[(a)] aligned with the computations of the target algorithm;
	\item[(b)] operates by matrix multiplications, hence natively admits useful gradients;
	\item[(c)] operates over high-dimensional latent spaces, hence is not vulnerable to bottleneck phenomena and may be more data-efficient.
\end{inparaenum}

\paragraph{The blueprint of neural algorithmic reasoning}
\begin{figure}
    \centering
    \includegraphics[width=\linewidth]{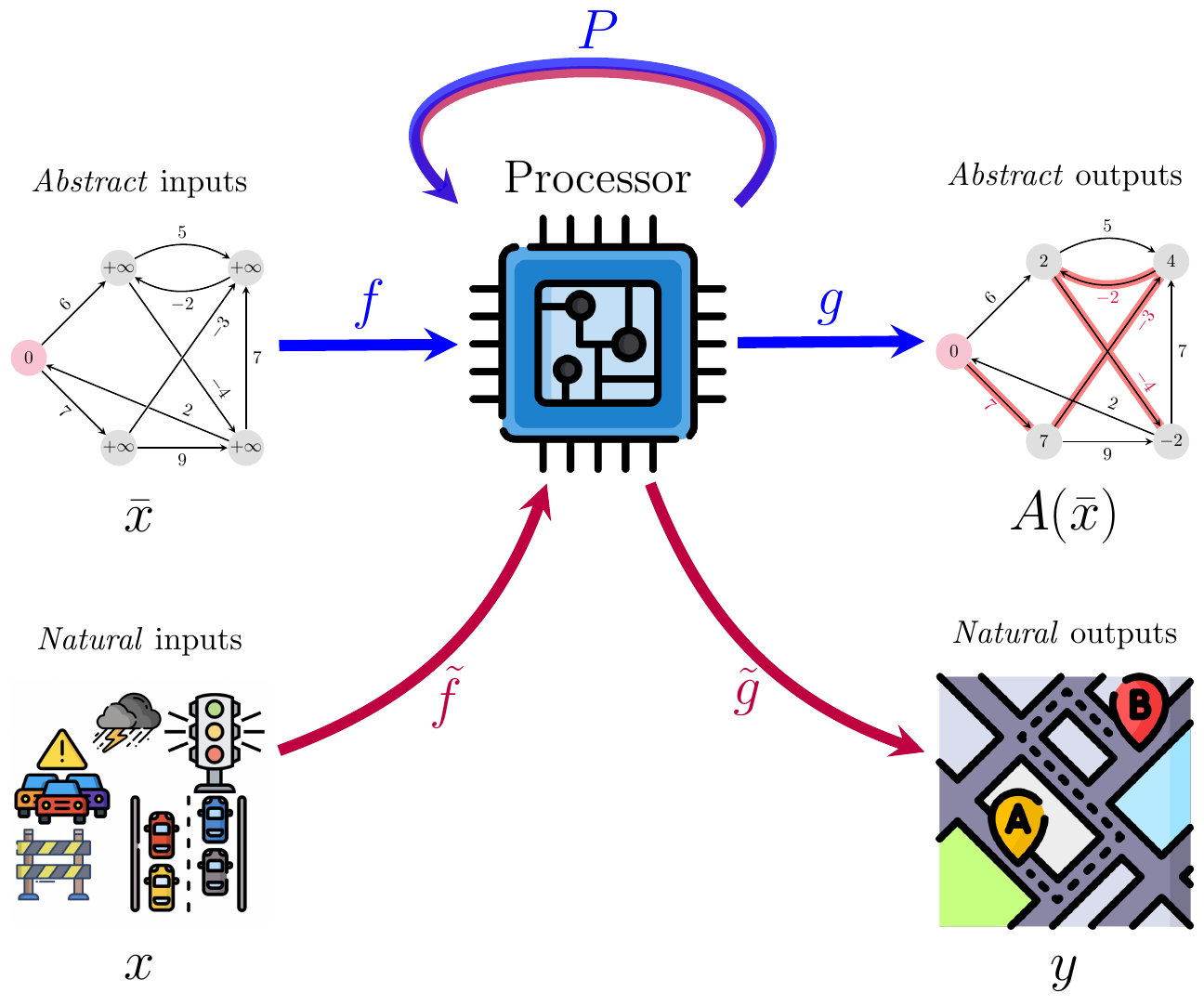}
    \caption{The blueprint of \emph{neural algorithmic reasoning}. We assume that our real-world problem requires learning a mapping from \emph{natural} inputs, $x$, to natural outputs, $y$---for example, fastest-time routing based on real-time traffic information. Note that natural inputs are often likely high-dimensional, noisy, and prone to changing rapidly---as is often the case in the traffic example. Further, we assume that solving our problem would benefit from applying an \emph{algorithm}, $A$---however, $A$ only operates over \emph{abstract} inputs, $\bar{x}$. In this case, $A$ could be \emph{Dijkstra}'s algorithm for shortest paths \citep{dijkstra1959note}, which operates over weighted graphs with exactly one scalar weight per node, producing the shortest path tree. First, an algorithmic reasoner is trained to imitate $A$, learning a function $g(P(f(\bar{x})))$, optimising it to be close to ground-truth abstract outputs, $A(\bar{x})$. $P$ is a \emph{processor network} operating in a high-dimensional latent space, which, if trained correctly, will be able to imitate the individual steps of $A$. $f$ and $g$ are encoder and decoder networks, respectively, designed to carry abstract data to and from $P$'s latent input space. Once trained, we can replace $f$ and $g$ with $\tilde{f}$ and $\tilde{g}$---encoders and decoders designed to process natural inputs into the latent space of $P$ and decode $P$'s representations into natural outputs, respectively. Keeping $P$'s parameters fixed, we can then learn a function $\tilde{g}(P(\tilde{f}(x)))$, allowing us an end-to-end differentiable function from $x$ to $y$, without any low-dimensional bottlenecks---hence it is a great target for neural network optimisation.}
    \label{fig:blueprint}
\end{figure}
Having motivated the use of \emph{neural algorithmic executors}, we can now demonstrate an elegant neural end-to-end pipeline which goes straight from raw inputs to general outputs, while emulating an algorithm internally. The general procedure for applying an algorithm $A$ (which admits abstract inputs $\bar{x}$) to raw inputs $x$ is as follows (following Figure \ref{fig:blueprint}):
\begin{enumerate}
	\item Learn an algorithmic reasoner for $A$, by learning to execute it on synthetically generated inputs, $\bar{x}$. This yields functions $f, P, g$ such that $g(P(f(\bar{x}))) \approx A(\bar{x})$. $f$ and $g$ are encoder/decoder functions, designed to carry data to and from the latent space of $P$ (the processor network).
	\item Set up appropriate encoder and decoder neural networks, $\tilde{f}$ and $\tilde{g}$, to process raw data and produce desirable outputs. The encoder should produce embeddings that correspond to the input dimension of $P$, while the decoder should operate over input embeddings that correspond to the output dimension of $P$.
	\item Swap out $f$ and $g$ for $\tilde{f}$ and $\tilde{g}$, and learn their parameters by gradient descent on any differentiable loss function that compares $\tilde{g}(P(\tilde{f}(x)))$ to ground-truth outputs, $y$. The parameters of $P$ should be kept \emph{frozen}.
\end{enumerate}
Through this pipeline, neural algorithmic reasoning offers a strong approach to applying algorithms on natural inputs. The raw encoder function, $\tilde{f}$, has the potential to replace the human feature engineer, as it is learning how to map raw inputs onto the algorithmic input space for $P$, purely by backpropagation.

One area where this blueprint had already proved useful is \emph{reinforcement learning} (RL). A very popular algorithm in this space is Value Iteration (VI)---it is able to solve the RL problem \emph{perfectly}, assuming access to environment-related inputs that are usually hidden. Hence it would be highly attractive to be able to apply VI over such environments, and also, given the partial observability of the inputs necessary to apply VI, it is a prime target for our reasoning blueprint. 

Specifically, the XLVIN architecture \citep{deac2020xlvin} is an exact instance of our blueprint for the VI algorithm. Besides improved data efficiency over more traditional approaches to RL, it also compared favourably against ATreeC \citep{farquhar2017treeqn}, which attempts to directly apply VI in a neural pipeline, thus encountering the algorithmic bottleneck problem in low-data regimes.

\paragraph{Conclusion}

We demonstrated how neural algorithmic reasoning can form a rich basis and core for learning novel and old algorithms alike.
At first, algorithmic reasoners can be bootstrapped from existing algorithms using supervision of their internal workings, and then subsequently embedded into the real world input/outputs via separately trained encoding/decoding networks. Such an approach has already proved fruitful across a range of domains, such as reinforcement learning and genome assembly. It is our belief that neural algorithmic reasoning will allow for applying classical algorithms on inputs that substantially generalise the preconditions specified by their designers, uniting the theoretical appeal of algorithms with their intended purpose.

\bibliography{main}
\bibliographystyle{plainnat}
\end{document}